\documentclass[10pt,twocolumn,letterpaper]{article}

\usepackage[margin=0.75in,columnsep=0.25in]{geometry}
\usepackage[T1]{fontenc}
\usepackage{times}
\usepackage{microtype}
\usepackage{amsmath}
\usepackage{booktabs}
\usepackage{array}
\usepackage{enumitem}
\usepackage{xcolor}
\usepackage{tikz}
\usetikzlibrary{positioning,arrows.meta,calc}
\usepackage[font=small,labelfont=bf]{caption}
\usepackage[hidelinks]{hyperref}
\usepackage{xurl}
\usepackage{tabularx}
\usepackage{multirow}
\usepackage{dblfloatfix}
\usepackage{balance}
\usepackage{titlesec}
\usepackage{needspace}
\titlespacing*{\paragraph}{0pt}{1.5ex plus 0.3ex minus 0.2ex}{0.8em}
\usepackage{pgfplots}
\pgfplotsset{compat=1.16}

\definecolor{cRaw}{HTML}{2A78D6}      
\definecolor{cGrepS}{HTML}{8DB8EC}    
\definecolor{cSum}{HTML}{EB6834}      
\definecolor{cMem}{HTML}{1BAF7A}      
\definecolor{cNone}{HTML}{8A8984}     
\colorlet{cGrep}{cMem}
\pgfplotsset{
  dotwhisker/.style={
    width=0.89\linewidth, xmin=0, xmax=1, clip=false, y dir=reverse, enlarge y limits={abs=0.6},
    xtick={0,0.2,0.4,0.6,0.8,1}, xmajorgrids, grid style={black!10},
    axis line style={black!45}, tick style={black!45}, ytick style={draw=none},
    tick label style={font=\footnotesize}, label style={font=\footnotesize},
    legend cell align=left,
    legend style={font=\scriptsize, draw=black!20, fill=white, inner sep=2pt, row sep=-1pt},
    error bars/error bar style={line width=1.1pt},
    error bars/error mark options={rotate=90, mark size=2.5pt, line width=1.1pt},
  },
  est/.style={only marks, mark size=2.6pt, line width=0.6pt, mark options={draw=white},
    error bars/x dir=both, error bars/x explicit},
  alt/.style={only marks, mark=diamond, mark size=2.6pt, line width=0.8pt, black!75},
  designbars/.style={
    ybar, width=\linewidth, height=5.4cm, ymin=0, ymax=118, bar width=4.2pt,
    symbolic x coords={Haiku,Sonnet,Luna,Sol}, xtick=data, enlarge x limits=0.16,
    ytick={0,20,40,60,80,100}, ymajorgrids, grid style={black!10},
    axis line style={black!45}, tick style={black!45}, xtick style={draw=none},
    tick label style={font=\footnotesize}, label style={font=\footnotesize},
    ylabel={Correct (\%)},
    nodes near coords, every node near coord/.append style={font=\tiny, rotate=90, anchor=west, black!80},
    legend cell align=left, legend columns=2,
    legend style={font=\scriptsize, draw=black!20, fill=white, inner sep=2pt, at={(0.5,1.02)}, anchor=south},
  },
  diffs/.style={
    width=0.89\linewidth, xmin=-40, xmax=90, y dir=reverse, enlarge y limits={abs=0.6},
    xtick={-40,-20,0,20,40,60,80}, xmajorgrids, grid style={black!10},
    axis line style={black!45}, tick style={black!45}, ytick style={draw=none},
    tick label style={font=\scriptsize}, label style={font=\footnotesize},
    error bars/error bar style={line width=1.1pt},
    error bars/error mark options={rotate=90, mark size=2.2pt, line width=1.1pt},
  },
}

\setlist{nosep,leftmargin=*}
\newcommand{\tool}[1]{\texttt{#1}}
\newcommand{\arm}[1]{\textsf{#1}}

\title{\Large\bfseries You Don't Need Long Context, You Need the Record}
\author{Adarsh Srivastava\\CtrlB Explore Inc.\\\texttt{adarsh@ctrlb.ai}}
\date{November 2026}

\begin{document}
\maketitle

\begin{abstract}
Frequent compaction keeps long-horizon agents accurate on task benchmarks while their context stays
short~\cite{nvshort}. We measure what that short context loses about the agent's own past. At real compaction
points in four engineers' coding-agent sessions, the post-compaction context held 6\% of exact answers and 40\% of
gist answers. Agents with only that context scored 22--35\% on questions about earlier work. On 50 questions, a
baseline given the history since the previous compaction scored 56\%, versus 90\% with search. With the full record in CtrlB
Recall and a one-line instruction to search it, four models in two families scored 81--94\%. Plain transcript files
searched with grep scored 77--94\%, lower on gist questions in three of the four models. In a synthetic scale test at
100 times the record, the grep calls that completed took minutes, where CtrlB stayed near one second. Without the instruction, agents often did
not search. CtrlB Recall also answered questions about the team's earlier sessions, which no summary answered.
With a searchable record, compaction can be lighter: removing the summary preserved observed recall in the tested
configurations, at lower cost.
\end{abstract}

\section{Introduction}
\label{sec:intro}

A coding agent such as Claude Code~\cite{claudecode} or Codex~\cite{codex} works in a loop. The model reads the
conversation, calls tools and appends their results~\cite{react}. When the context window fills, both harnesses
\emph{compact} by default: a model writes a summary, and the agent continues from it. Whatever the summary omits is
no longer in the agent's context.

Recent work argues that long contexts are not needed. Compacting at total context limits of 8k--48k tokens kept
task accuracy close to full history on software-engineering, terminal, search and tool-use benchmarks, and on
SWE-bench Verified and Terminal-Bench even compacting after every action came within a few
points~\cite{nvshort}. That work gives a constructed example of the limit: an agent that must recall an arbitrary
old coin toss will eventually lose it, unless the harness keeps every message and lets the agent read it
selectively. It does not measure how much real compactions lose, or whether agents use such a record.

We build that record and measure both. In a \emph{session store}, every session's full record is kept and
searchable, and a one-line standing rule tells the agent to search it before answering about earlier work. We compare it with compaction alone, with
Mem0~\cite{mem0paper} (a memory store that extracts facts), and with keeping everything in the window, on two model
families and two model sizes.

We implement the session store with CtrlB Recall, built by the author's company, and compare it with plain
transcript files that the agent searches with \tool{grep}. Behind the same recall tool, CtrlB and a \tool{grep} backend returned identical
rows on every replayed call (Section~\ref{sec:system}), so the recall gains come from keeping the record and from
the tool, not from a different match rule. Against raw transcript files, the tool's rows raised gist accuracy
(Section~\ref{sec:rule}), and in a synthetic scale test at 100 times the record, CtrlB's index kept search near one
second, where the \tool{grep} calls that completed took minutes (Section~\ref{sec:speed}).

\Needspace{4\baselineskip}
\paragraph{Findings.}
\begin{itemize}
  \item \textbf{Compaction retained more current-state answers than exact answers.} The post-compaction context held 6\% of
        exact answers and 40\% of gist answers, and agents given only that context scored 22--35\% on Detail
        recall. Larger models still failed to recover most missing details (Sol 35\%, Luna 29\%).
  \item \textbf{A longer window recovered fewer details than search.} On 50 questions, Haiku with a median of 186k tokens of
        context scored 56\%, against 90\% with search. In 10 of the 29 exact questions the answer predated the
        window; where it was in the window, full context found 12 of 19 and search 17 of 19.
  \item \textbf{Search raised Detail recall to 81--94\% in every model tested.} CtrlB matched or
        beat grep over the transcript file in every model, by up to 6 points overall and by 13--14 points on gist
        questions for the Claude models. Small models with the record found the next step and its outcome in
        70--72\% of lookups in the team's earlier sessions; a summary answered none.
  \item \textbf{The agent must be told to search.} Without the standing rule, summary-equipped agents made no
        search on 48--89\% of gist questions. The rule raised Detail recall by 5--47 points.
  \item \textbf{Removing the summary preserved observed recall in the tested configurations,} at the lowest answer
        cost, though some intervals allow meaningful losses. Background running notes achieved first-action scores of 100\%
        (Haiku) and 97\% (Luna), versus 95\% with compaction. Their on-task scores were 11 and 5 points lower, not
        significant on 38 points. They cost \$0.008--0.013 per point, against \$0.08--0.23 and a median
        61-second pause for cached compaction.
  \item \textbf{Small models with the record beat large models without it} by 37--63 points on Detail recall,
        against summaries or Mem0. With the store on both sides, Luna trailed Sol by 8--13 points and Haiku's
        differences from Sonnet were inconclusive.
\end{itemize}

\section{Background and related work}
\label{sec:related}

\paragraph{Long contexts and compaction.}
Models use information in the middle of long contexts less reliably than at the ends~\cite{lostmiddle}, and cost
grows with context length. Harnesses therefore bound the window. Claude Code
compacts automatically near the limit and on the \tool{/compact} command~\cite{claudecode}. Its summary is a user
message that ends, in the versions we study, with the path of the session's JSON Lines (JSONL) transcript. The path
is present in all 72 Claude Code compactions in our corpus. Codex writes a compaction item into its own session
file; none of the 32 in our corpus mentions where the record is.

\paragraph{Short contexts.}
Belcak et al.~\cite{nvshort} show that compacting at total context limits of 8k--48k tokens (thresholds of
6k--36k) keeps the accuracy of Claude Opus~4.5 and GPT~5.6 Sol within a few points of full history on SWE-bench
Verified and Multilingual, Terminal-Bench, FRAMES, $\tau^2$-bench and BFCL. Reading long inputs selectively matches
or beats ingesting them whole, and compacting after every action stays within a few points on SWE-bench Verified
and Terminal-Bench. Their best compactions are briefs for the work ahead (task, current activity, plan,
requirements), not narratives of the trace. Our survival measurements are consistent with that design: Claude
Code's compactions kept the current state in 60\% of cases and commands, errors and numbers in 0--4\%. One
hypothesis consistent with both studies is that their benchmarks rarely need an old exact detail. We did not test
their compactions or tasks, and their benchmarks score completed tasks where ours ask about the past. Their
single-action regime carries a natural-language state between calls; our running note is a related idea.

\paragraph{Memory and retrieval.}
MemGPT~\cite{memgpt} pages information between the context and external storage. Mem0~\cite{mem0paper} and the
memory stream of generative agents~\cite{genagents} distil turns into short facts with embeddings. LoCoMo~\cite{locomo}
and LongMemEval~\cite{longmemeval} evaluate long-term memory in multi-session conversations. Coding-agent sessions
differ: most of their bytes are tool output, and many questions need an exact string back. Coding-agent benchmarks
measure task completion within a session; we know of none that asks about a session's past after compaction, or
about other sessions. Retrieval-augmented generation~\cite{rag} usually retrieves from a static corpus, while here
the corpus is the agent's own growing history. The agent--computer interface strongly affects agent
performance~\cite{sweagent}: here the tool's shape changed accuracy, and whether the agent called it at all
depended on a single instruction. The Model Context Protocol (MCP)~\cite{mcp} lets a harness attach a tool without
changing it.

\section{System: the session store}
\label{sec:system}

\begin{figure}[t]
\centering
\begin{tikzpicture}[
  box/.style={draw, rounded corners=2pt, align=center, font=\footnotesize, minimum height=0.9cm, inner sep=3pt},
  arr/.style={-{Stealth[length=2mm]}, thick}]
  \node[box, minimum width=2.6cm] (agent) {Agent harness\\(model + standing rule)};
  \node[box, minimum width=2.6cm, right=1.1cm of agent] (rec) {Session files\\(Claude Code, Codex)};
  \node[box, minimum width=6.3cm, below=0.9cm of $(agent)!0.5!(rec)$] (store)
       {Session store: one row per content block\\ CtrlB Recall \emph{or} JSONL files with grep};
  \node[box, minimum width=2.6cm, below=0.9cm of store.south west, anchor=north west] (tool)
       {Recall tool (MCP)\\search $\cdot$ read $\cdot$ what ran next};
  \draw[arr] (agent) -- node[above, font=\scriptsize]{every step} (rec);
  \draw[arr] (rec.south) -- (rec.south |- store.north);
  \draw[arr] (store.south -| tool.north) -- (tool.north);
  \draw[arr] (tool.west) -- ++(-0.35,0) |- node[pos=0.25, left, font=\scriptsize, align=right]{short\\replies} (agent.west);
\end{tikzpicture}
\caption{The session store. Every step is appended to the session file; a normaliser turns it into rows; the
agent reads them through one tool, and a standing rule in its system prompt tells it to search before answering
about earlier work. When the window fills, the harness compacts as usual or starts a fresh context that points to
the store; either way the full record stays searchable.}
\label{fig:arch}
\end{figure}
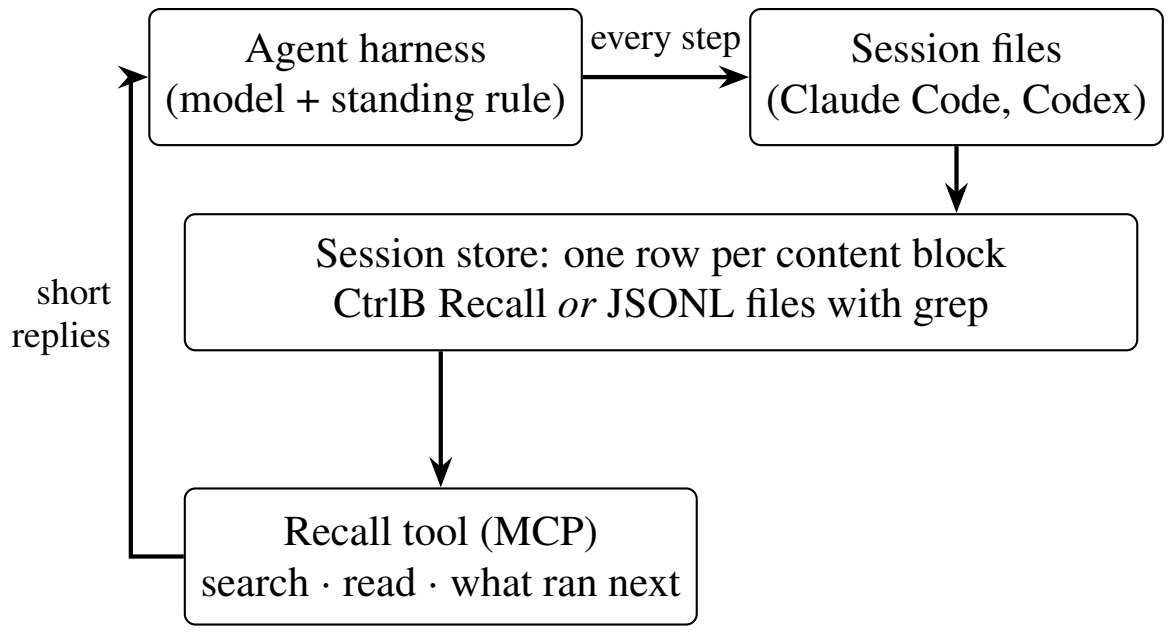

\paragraph{Rows.}
Figure~\ref{fig:arch} shows the design. We normalise both harnesses' files into one row per content block: text, thinking, tool call and tool
result blocks in Claude Code; messages, reasoning summaries, function and custom tool calls, and their outputs in
Codex. Each row records the session, the line number in the file (\emph{seq}), the block index, role,
block type, tool name, shell command, error flag, summary flag, working directory, timestamp and body. A failure is
the tool result's error flag in Claude Code, or a nonzero exit code in Codex. For a tool call, the body is the
shell command (or the input JSON), so commands are searchable.

\paragraph{Recall tool.}
The first version has three operations: \textbf{search} (rows matching a term, with session, seq, kind and a
300-character excerpt), \textbf{read} (the rows of one session in a seq range, at most 50, bodies cut at 4,000
characters) and \textbf{what followed} (for each tool call or result matching a term, the next records of its
session). The tool can be scoped to one session and to rows before a compaction point, so an agent cannot see the
future. Section~\ref{sec:lookupfix} describes a second version for Cross-session lookup, designed from the first
version's failure traces: \tool{find\_command} (every past call whose command contains a string, newest first,
with a total and paging), \tool{search\_sessions} with order, paging and one hit per session,
\tool{next\_actions} (from a call, its result and the next $k$ tool calls with their results, skipping text) and
\tool{fetch\_records}; every reply is capped at 12,000 characters.

\paragraph{Two backends.}
\textbf{CtrlB Recall} (built by the author's company) stores rows as Apache Parquet~\cite{parquet} with sidecar index files (Iceberg Puffin~\cite{puffin}
and its own formats); the tool uses its exact-token search and Structured Query Language (SQL) queries.
\textbf{grep} keeps the same rows as one JSONL file per session and scans them with \tool{grep -F}. Both apply one
matching rule, taken from CtrlB's tokenizer: a token is a run of letters, digits, `.' and `+', only tokens of
3--50 bytes are indexed, and a term matches when it occurs case-insensitively with no token character directly
before or after it. On 98 test calls the two backends returned byte-identical replies, and on 1,352 calls replayed
from the Codex runs, identical rows. Identical retrieval results isolate backend latency from retrieval correctness; agent-level accuracy can still vary between runs.

\paragraph{Standing rule.}
Appended to the system prompt (Claude Code: \tool{--append-system-prompt}; Codex: developer instructions):
``Your earlier turns in this session are not all in your context: the compaction summary leaves details out.
[Where the record is.] Before answering anything about earlier work in this session, search it. Do not say you
cannot find something until you have searched.'' The bracket names the transcript file or the recall tools. For
Cross-session lookup the rule says the team's earlier sessions are not in context and should be searched.

\section{Experimental setup}
\label{sec:setup}

\paragraph{Corpus.}
Table~\ref{tab:corpus} describes the corpus: one-time exports of four engineers' sessions between 18 September and
31 October 2026. Three engineers used Claude Code and one used Codex. The exporter redacted teammates' exports. The
corpus holds proprietary code and is not released, and no session content appears in this paper. All 72 compaction
points are in Claude Code sessions, so both model families answer questions about Claude Code sessions. Codex
sessions enter Cross-session lookup, the store and the speed measurements.

\begin{table}[t]
\centering\small
\caption{Corpus and store. Eng.\ = engineer.}
\label{tab:corpus}
\setlength{\tabcolsep}{3.5pt}
\begin{tabular}{@{}lrrrrr@{}}
\toprule
 & Eng.\,1 & Eng.\,2 & Eng.\,3 & Eng.\,4 & Total \\
\midrule
Harness & Claude & Claude & Claude & Codex & \\
Session files & 537 & 223 & 193 & 116 & 1,069 \\
Raw MB & 847 & 422 & 862 & 104 & 2,236 \\
Rows (k) & 70.5 & 45.8 & 48.7 & 5.4 & 170.4 \\
Shell calls (k) & 19.1 & 12.9 & 12.7 & 1.3 & 46.0 \\
Compaction points & 33 & 10 & 29 & 0 & 72 \\
\bottomrule
\end{tabular}
\vspace{2pt}

\footnotesize CtrlB: 36,194 rows/s ingest; 169\,MB of Parquet and index files for 263\,MB of row text.
\end{table}

\paragraph{Tasks.}
\begin{itemize}
  \item \textbf{Detail recall.} At a real compaction point, the agent is asked about the session's history.
        \emph{Exact} questions (174 in the full set) ask for a path, number, error line or command from a tool
        result or shell command before the point, occurring 1--3 times there, sampled \emph{without looking at
        the summary} across three distance bands. Haiku~5.5 phrased each question; we dropped 2 that contained
        their answer and 31 that spelled out most of its words. \emph{Gist} questions (135) ask for a decision,
        result, current state, reason or abandoned approach; Sonnet~5.5 wrote each, with a one- or two-sentence
        reference answer, from a window of about 24k characters of the record, never seeing the summary. Exact
        answers are correct when the whitespace-normalised gold occurs in them; gist answers are judged against
        the reference.
  \item \textbf{Cross-session lookup.} ``Earlier, in one of the team's coding-agent sessions, running \emph{X}
        failed with \emph{Y}. What did the agent do next, and did it work?'', with no session given. The outcome
        is the error flag of the next shell call within eight records, or ``no retry''. Each (command, error)
        pair occurs in exactly one episode. The full set has 188 questions (130 fixed, 41 still failing, 17 no
        retry) from 146 sessions, 36 of them Codex. A judge decides whether the answer states the same next step
        and outcome (\emph{correct}); \emph{exact next} checks that the next command is quoted verbatim.
  \item \textbf{Resumption.} At the 38 points followed by a real user message and at least three tool calls, the
        agent writes a plan (task, first three concrete actions, facts relied on) without running anything,
        because the repositories have changed since. A judge decides whether one of the three actions is
        substantially one of the real agent's first five (\emph{first action}), whether the plan is on the same
        task, and whether it states something the record contradicts.
\end{itemize}

\paragraph{The summary arms see what the harness shows.}
For every point we rebuild the context the agent had right after compaction from the session file: the summary,
the messages the harness kept verbatim (listed in the compaction boundary record) and the re-attached context
(file contents, plan file, task status, names of referenced files); the median is 29k characters. Where an arm
has the transcript, the summary's path names a copy of the session file cut at the compaction point.

\paragraph{Arms.}
Table~\ref{tab:arms} lists the memory designs. Each is run with a small and a large model of each family.

\begin{table}[t]
\centering\small
\caption{Designs compared. Summary-only, Mem0, default-harness and summary-plus-store arms start from the post-compaction context. The remaining arms vary or replace that context.}
\label{tab:arms}
\begin{tabularx}{\linewidth}{@{}>{\raggedright\arraybackslash}p{2.45cm}>{\raggedright\arraybackslash}X@{}}
\toprule
Design & What the agent gets \\
\midrule
\arm{Summary only} & The post-compaction context; no tools. Pure compaction. \\
\arm{Memory store (Mem0)} & Plus a \tool{search\_memories} tool over Mem0 memories extracted from the record before
  the point, and the standing rule. \\
\arm{Default harness} & Plus the transcript path at the end of the summary and a read-only shell, as stock
  Claude Code; no rule. (Codex does not supply the path; we add it.) \\
\arm{Session store (grep)} & The default harness plus the standing rule: the agent searches the transcript file
  with its shell. \\
\arm{Session store (CtrlB)} & Plus the recall tool on CtrlB, scoped to the session before the point, and the rule;
  no shell. \\
\arm{Record only + recall} & No summary: ``You are continuing session $X$\ldots{} retrieve what you need from
  its record with the recall tools'', the records since the cut and the user's message; recall on CtrlB. Sonnet,
  first study. \\
\arm{Running note} & No compaction summary: a note that Haiku or Luna updated as the session went, from the record
  in parts of at most 120k characters with tool outputs cut short, citing record positions; plus the kept messages,
  re-attached files and a pointer to the recall tools; the rule. \\
\arm{Hand-off} & As the running note, without the note: kept messages, re-attached files, pointer, rule. \\
\arm{Pointer only} & ``You are continuing session $X$\ldots'' and the rule; nothing else. \\
\arm{Full context} & Everything since the previous compaction, and that compaction's summary, in the prompt (bodies capped at 30,000 characters);
  no tools. \\
\midrule
\multicolumn{2}{@{}l}{\emph{Cross-session lookup, no session given}} \\
\arm{Summary only} & The question alone; the agent has no other sessions. \\
\arm{Shell, no rule} & A read-only shell over every session file of the four engineers. \\
\arm{Session store (grep / CtrlB)} & The same files with the rule, or the recall tool over the whole corpus with
  the rule. \\
\bottomrule
\end{tabularx}
\end{table}

\paragraph{Mem0 store.}
We run open-source Mem0 2.2.1~\cite{mem0paper} with all-MiniLM embeddings and a FAISS index; its extraction calls
go to gpt-6-luna. Each Detail point gets its own store, built only from the record before the point, so the agent
never sees memories from after its point. The Lookup store is \emph{favourable}: it holds the 60 questions' source
windows plus 120 distractor windows, where a real team memory would hold every session. It favours Mem0 on lookup
but does not establish a formal upper bound for other deployments. The reader is Luna, or Sol reading the same
Luna-written memories. Build sizes and costs are in Appendix~\ref{app:details}.

\paragraph{Models, runners and samples.}
Claude arms run through the Claude Code command-line interface (CLI) in headless mode (\tool{claude -p}) with
Haiku~5.5 or Sonnet~5.5, session persistence off and a 20-turn limit. Codex arms run through \tool{codex exec}
(version 0.164.0, reasoning effort medium) with gpt-6-luna or gpt-6.1-sol (and gpt-6-astra as an extra reference),
in a read-only sandbox inside a private mount namespace that shows each run only its allowed files. List prices per
million tokens (input/output) are \$0.10/\$0.50 for Haiku and Luna and \$2/\$10 for Sonnet and Sol, so the small
models are 20 times cheaper per token in both families. Claude costs are the CLI's reported
\tool{total\_cost\_usd}. Codex costs are estimated from token counts at list prices and are not subscription
charges.

The memory-design and model-size comparisons use 100 Detail questions per family drawn from the full 309 with
session and type strata (35 sessions per family; Appendix~\ref{app:details}), and the same 60 Cross-session lookup questions in both families. A further 60 held-out
lookup questions test the redesigned tool. Section~\ref{sec:nocompact} reports a first study on the full 309, 188 and
38 questions with Sonnet and, without the rule, with Luna, Sol and Astra.

\paragraph{Judges.}
To avoid a model grading its own family, gpt-6-astra judges every Claude arm in the comparison and the
lighter-compaction runs (Section~\ref{sec:replace}). gpt-6.1-sol, blinded to arm and model, judges the other Codex
arms. Sol therefore grades some answers Sol wrote; a seeded 20\% sample re-judged by Astra checks agreement but
does not rule out bias. The first study and the full-context arm use a Sonnet judge. Call counts and agreement
rates are in Appendix~\ref{app:details}.

\paragraph{Statistics.}
Questions from one session are not independent, so 95\% confidence intervals (CI) resample sessions (10,000 draws;
2,000 in the first study)~\cite{bootstrap}, and differences are paired on the same questions. We call a smaller
model \emph{non-inferior} when the interval's lower end is above $-5$ points. Exact McNemar tests~\cite{mcnemar}
are used in the first study.

\paragraph{File access.}
Claude's shell arms can Grep or Read any path on the machine. In 12 Detail and 2 lookup questions a shell-arm run
read live files outside its folder. Dropping those questions from every arm changes no conclusion
(Appendix~\ref{app:details}). Codex runs saw only their allowed files.

\section{Results}
\label{sec:results}

\subsection{What compaction keeps}
\label{sec:keeps}
We first measured the context itself on the full 309 questions. The summary text contained 2 of the 174 exact golds
(1\%). With the verbatim messages and re-attached files, the post-compaction context contained 11 (6\%), and it
stated the answer to 54 of 135 gist questions (40\%). Survival by distance and kind is in Appendix~\ref{app:survival}. The retained context preserved current state
(60\% survival) more often than commands and error lines (0\%), numbers (4\%) and the other details that led to
it.

\subsection{The store with and without the summary}
\label{sec:nocompact}

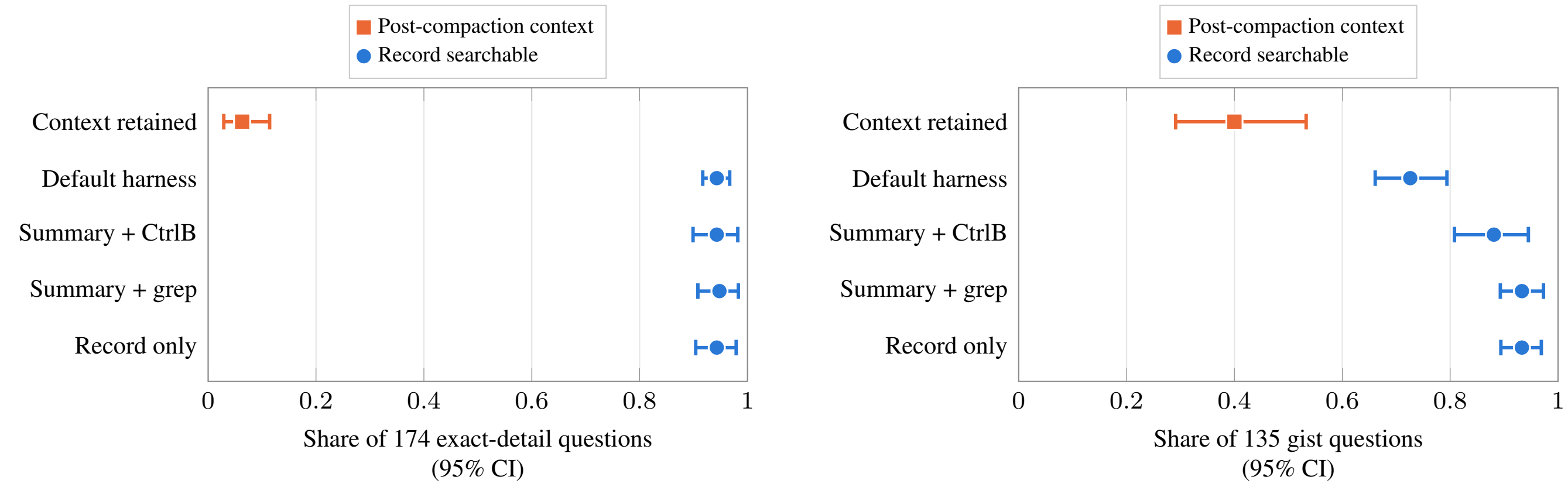
\begin{figure*}[!t]
\centering
\begin{minipage}[t]{0.485\textwidth}
\centering
\begin{tikzpicture}
\begin{axis}[dotwhisker, height=4.9cm,
  ytick={1,2,3,4,5}, yticklabels={Context retained, Default harness, Summary + CtrlB, Summary + grep, Record only},
  xlabel={Share of 174 exact-detail questions\\(95\% CI)}, xlabel style={align=center},
  legend style={at={(0.5,1.03)}, anchor=south, legend columns=1}]
\addplot+[est, mark=square*, color=cSum] coordinates {(0.063,1) += (0.051,0) -= (0.034,0)};
\addlegendentry{Post-compaction context}
\addplot+[est, mark=*, color=cRaw] coordinates {(0.943,2) += (0.024,0) -= (0.026,0) (0.943,3) += (0.039,0) -= (0.044,0)
  (0.948,4) += (0.035,0) -= (0.040,0) (0.943,5) += (0.036,0) -= (0.039,0)};
\addlegendentry{Record searchable}
\end{axis}
\end{tikzpicture}

\small (a) Exact details: path, number, error line, command
\end{minipage}\hfill
\begin{minipage}[t]{0.485\textwidth}
\centering
\begin{tikzpicture}
\begin{axis}[dotwhisker, height=4.9cm,
  ytick={1,2,3,4,5}, yticklabels={Context retained, Default harness, Summary + CtrlB, Summary + grep, Record only},
  xlabel={Share of 135 gist questions\\(95\% CI)}, xlabel style={align=center},
  legend style={at={(0.5,1.03)}, anchor=south, legend columns=1}]
\addplot+[est, mark=square*, color=cSum] coordinates {(0.400,1) += (0.133,0) -= (0.109,0)};
\addlegendentry{Post-compaction context}
\addplot+[est, mark=*, color=cRaw] coordinates {(0.726,2) += (0.068,0) -= (0.065,0) (0.881,3) += (0.064,0) -= (0.073,0)
  (0.933,4) += (0.040,0) -= (0.040,0) (0.933,5) += (0.036,0) -= (0.039,0)};
\addlegendentry{Record searchable}
\end{axis}
\end{tikzpicture}

\small (b) Gist: decisions, results, state, reasons
\end{minipage}
\caption{Detail recall with Sonnet~5.5 at 70 compaction points, no standing rule (309 questions). Squares: share
of questions whose answer is still in the context after compaction. Circles: agents that can search the record
before the point. The agent with no summary at all (bottom row) achieves similar observed accuracy to the summary-equipped recall arms, on exact
details and gist; the default harness, which compacts and points to the transcript, trails on gist
($p<0.001$).}
\label{fig:detail}
\end{figure*}

Our first study tested whether the summary adds recall when the record is searchable. It ran Sonnet on the full
309 Detail, 38 Resumption and 188 lookup questions, without the standing rule, and compared the stock harness
(compaction plus the transcript path and a shell), compaction plus the recall tool, and the record with \emph{no
summary}. In the last arm the agent starts from ``You are continuing session $X$. Its earlier turns are no longer
in your context. Retrieve what you need from its record with the recall tools as you go'', the records since the
cut and the user's message (the design of the Mainplane harness~\cite{mainplane}). The agent with no summary answered 0.94 of the 309 questions,
against 0.92 and 0.94 for the two arms with the summary and the same tool, and 0.85 for the stock harness
(Figure~\ref{fig:detail}). We found no recall advantage from the summary in this experiment. The agent with no
summary searched in every gist run, and it was the cheapest per question. Table~\ref{tab:nocompact} and the detailed
comparisons are in Appendix~\ref{app:first}.

\subsection{Resumption and task orientation}
\label{sec:resume}

\begin{figure}[!htb]
\centering
\begin{tikzpicture}
\begin{axis}[dotwhisker, height=4.6cm,
  ytick={1,2,3,4}, yticklabels={Message only, Default harness, Summary + recall, Record only},
  xlabel={Share of 38 plans\\(95\% CI for first action)}, xlabel style={align=center},
  legend style={at={(0.5,1.03)}, anchor=south, legend columns=2}]
\addplot+[est, mark=pentagon*, color=cNone] coordinates {(0.763,1) += (0.130,0) -= (0.096,0)};
\addlegendentry{No memory (control)}
\addplot+[est, mark=square*, color=cSum] coordinates {(1.0,2) += (0,0) -= (0.092,0) (1.0,3) += (0,0) -= (0.092,0)};
\addlegendentry{With summary}
\addplot+[est, mark=*, color=cRaw] coordinates {(0.842,4) += (0.102,0) -= (0.098,0)};
\addlegendentry{Record, no summary}
\addplot[alt] coordinates {(0.447,1) (0.368,2) (0.237,3) (0.368,4)};
\addlegendentry{Contradiction rate}
\end{axis}
\end{tikzpicture}
\caption{Resumption at 38 real compaction points (Sonnet). Filled marks: one of the plan's first three actions
matches the real agent's first five. Diamonds: plans stating something the record contradicts (lower is better).
Intervals for the two 1.00 arms are Wilson bounds.}
\label{fig:resume}
\end{figure}
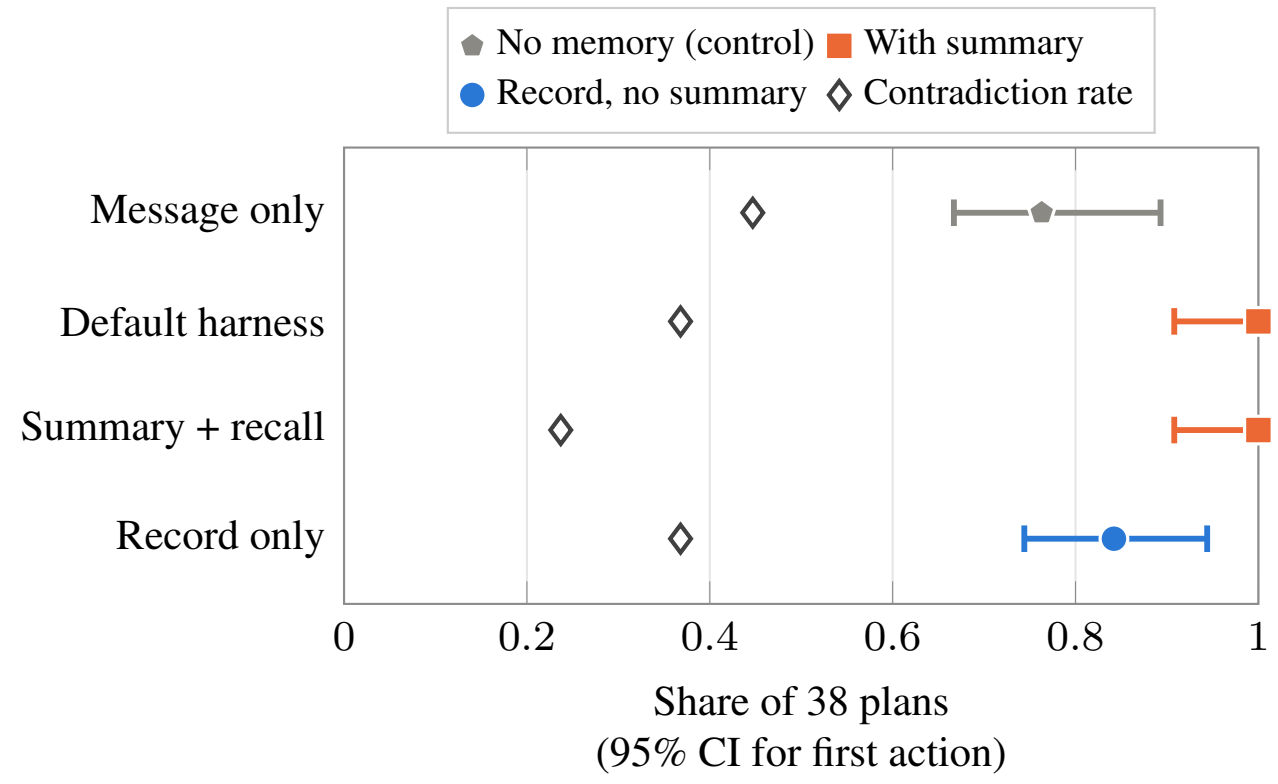

The summary improved task resumption in the first study (Figure~\ref{fig:resume}). With the post-compaction
context, every Sonnet plan contained the real first action (38 of 38, with or without the recall tool). With the
record and no summary, 32 of 38 did ($p=0.03$, 6 vs 0 discordant); with neither, 29 of 38. The record-only plans
named fewer of the identifiers the real agent went on to use (0.22 vs 0.31) and took longer (24.8 vs 16--18\,s),
because the agent made a median of 4.5 recall calls first. Their contradiction rate equalled the stock harness's
(0.37). On Codex, with the summary, Luna, Sol and Astra all found the first action at all 38 points.

Two points temper this. In an earlier pilot at 21 points, the summary arm received the summary alone and the
record-only arm tied it (0.95 each); the 16-point gap appeared once the summary arm also got the verbatim recent
messages and re-attached files that Claude Code keeps without a separate model call. Also, six plans in 38 is the whole
difference. Section~\ref{sec:replace} tests lighter alternatives.

\subsection{What compaction costs}
\label{sec:cost}
Compaction is not free. The 72 Claude Code compactions in the corpus record their own duration. The agent stopped
for a median of 61\,s (90th percentile 129\,s, longest 206\,s), at a median of 336k tokens of context, and 37 of
the compactions were triggered automatically. A cached re-summarisation cost \$0.08--0.23 per point
(Appendix~\ref{app:cost}). This pause and cost are substantial. The recall experiments found limited retention of
exact details by the summary and no detected recall advantage from it when the record was searchable.

\subsection{Lighter compaction once the record is kept}
\label{sec:replace}

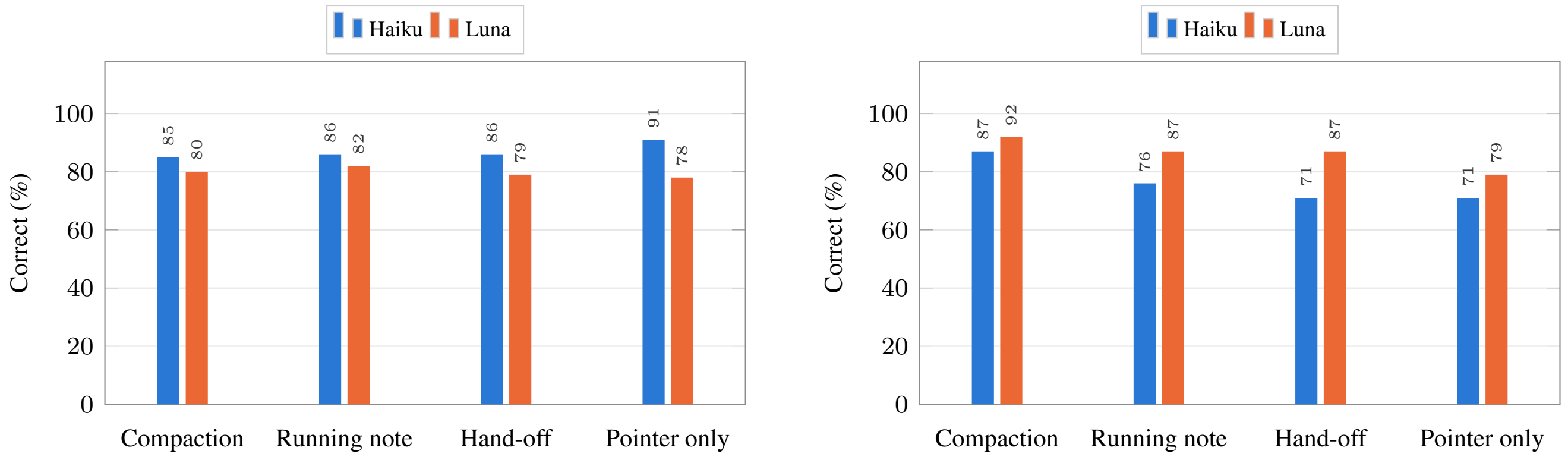
\begin{figure*}[!t]
\centering
\begin{minipage}[t]{0.49\textwidth}
\centering
\begin{tikzpicture}
\begin{axis}[designbars, symbolic x coords={Compaction,Running note,Hand-off,Pointer only},
  enlarge x limits=0.16, bar width=7pt, legend columns=2]
\addplot[fill=cRaw, draw=none] coordinates {(Compaction,85) (Running note,86) (Hand-off,86) (Pointer only,91)};
\addlegendentry{Haiku}
\addplot[fill=cSum, draw=none] coordinates {(Compaction,80) (Running note,82) (Hand-off,79) (Pointer only,78)};
\addlegendentry{Luna}
\end{axis}
\end{tikzpicture}

\small (a) Detail recall: share correct (100 questions)
\end{minipage}\hfill
\begin{minipage}[t]{0.49\textwidth}
\centering
\begin{tikzpicture}
\begin{axis}[designbars, symbolic x coords={Compaction,Running note,Hand-off,Pointer only},
  enlarge x limits=0.16, bar width=7pt, legend columns=2]
\addplot[fill=cRaw, draw=none] coordinates {(Compaction,87) (Running note,76) (Hand-off,71) (Pointer only,71)};
\addlegendentry{Haiku}
\addplot[fill=cSum, draw=none] coordinates {(Compaction,92) (Running note,87) (Hand-off,87) (Pointer only,79)};
\addlegendentry{Luna}
\end{axis}
\end{tikzpicture}

\small (b) Resumption: plans on the same task as the real agent (38 points)
\end{minipage}
\caption{What the agent is shown when the window fills, once the record is kept, with the small model of each
family. Every arm has the CtrlB recall
tool scoped to the session before the point and the standing rule. \emph{Compaction}: Claude Code's
post-compaction context. \emph{Running note}: a note the same small model updated in the background, plus the kept
messages and re-attached files. \emph{Hand-off}: the kept messages and files only. \emph{Pointer only}: nothing but
where the record is. Recall point estimates are similar across the options, while on-task resumption scores are lower than with compaction. The running note gives the closest observed on-task score in each family. Judge: gpt-6-astra for both families.}
\label{fig:replace}
\end{figure*}

\begin{table*}[!t]
\centering\small
\caption{Lighter compaction: Haiku (Claude) and Luna (Codex), each with recall and the standing rule. Detail
recall: 100 questions per family (exact / gist in brackets). Resumption: 38 points. Differences against compaction
in points with 95\% CI, paired on the same questions. Judge: gpt-6-astra. Cost per question for answering; notes
cost \$0.013 (Haiku) and \$0.008 (Luna) per compaction point to write.}
\label{tab:replace}
\setlength{\tabcolsep}{3.4pt}
\begin{tabular}{@{}llcccccc@{}}
\toprule
 & & Detail correct & vs compaction & First action & Stayed on task & vs compaction & Contradicted \\
\midrule
\multirow{4}{*}{Haiku}
 & Compaction   & 85 (91 / 78) & ---                 & 95  & 87 & ---                  & 47 \\
 & Running note & 86 (93 / 78) & $+1$ ($-7$, $+10$)  & 100 & 76 & $-11$ ($-26$, $0$)   & 37 \\
 & Hand-off     & 86 (95 / 76) & $+1$ ($-5$, $+7$)   & 95  & 71 & $-16$ ($-30$, $-3$)  & 42 \\
 & Pointer only & 91 (98 / 82) & $+6$ ($-2$, $+16$)  & 89  & 71 & $-16$ ($-32$, $-3$)  & 42 \\
\midrule
\multirow{4}{*}{Luna}
 & Compaction   & 80 (83 / 76) & ---                 & 95  & 92 & ---                  & 24 \\
 & Running note & 82 (87 / 76) & $+2$ ($-7$, $+12$)  & 97  & 87 & $-5$ ($-22$, $+6$)   & 24 \\
 & Hand-off     & 79 (89 / 67) & $-1$ ($-8$, $+7$)   & 95  & 87 & $-5$ ($-24$, $+7$)   & 5 \\
 & Pointer only & 78 (78 / 78) & $-2$ ($-10$, $+8$)  & 97  & 79 & $-13$ ($-31$, $0$)   & 16 \\
\midrule
\multicolumn{2}{@{}l}{\$ / question, Haiku} & \multicolumn{6}{l}{Compaction 0.0054; running note 0.0045; hand-off 0.0041; pointer only 0.0035} \\
\multicolumn{2}{@{}l}{\$ / question, Luna}  & \multicolumn{6}{l}{Compaction 0.0027; running note 0.0023; hand-off 0.0023; pointer only 0.0018} \\
\bottomrule
\end{tabular}
\end{table*}

We then ran the small model of each family, Haiku and Luna, with three lighter alternatives to the compaction
summary, each with the recall tool and the standing rule (Figure~\ref{fig:replace}, Table~\ref{tab:replace}). The
\emph{running note} is a summary written while the session runs, not at the limit. The same small model reads the
session's record in parts of at most 120k characters, with tool outputs cut to 600 characters, and updates a note
of at most 500 words. The note holds the task, the current state, the next step, and the key decisions, results,
abandoned approaches and files, each citing the record positions that show it (Appendix~\ref{app:prompts}).
\begin{itemize}
  \item \textbf{Observed recall remained similar.} On Detail recall every alternative was within $-2$ to $+6$
        points of compaction in both families, none significantly different. The intervals do not establish
        equivalence or non-inferiority for every arm. Haiku with only a pointer scored highest (91\% against 85\%);
        on Luna the four arms were within 4 points of each other. Every alternative was cheaper per answer than
        compaction, the pointer-only arm by a third, because no summary is carried into each call.
  \item \textbf{The running note gave the orientation that compaction gave.} It found the real first action in
        100\% (Haiku) and 97\% (Luna) of plans, against 95\% for compaction in both. Its plans contradicted the
        record no more often (37\% vs 47\%; 24\% vs 24\%). They stayed on the real task slightly less often, by 11
        points on Haiku and 5 on Luna, neither significant at 38 points.
  \item \textbf{Recent messages and orientation notes improved task resumption.} With only a pointer, plans stayed
        on task 13--16 points less often than with compaction. Haiku made no recall call in 82\% of pointer-only
        plans, despite the rule, and planned from what it was shown. Luna searched (8.1 calls per plan) and
        recovered more. The kept messages and files alone were enough for Luna (on task $-5$, and the fewest
        contradictions, 5\%) but not for Haiku ($-16$).
  \item \textbf{Background note updates reduced writing cost and avoided synchronous summarisation.} The notes cost
        \$0.008--0.013 per compaction point, against \$0.08--0.23 and a stop of about a minute for cached
        compaction (Appendix~\ref{app:cost}).
\end{itemize}

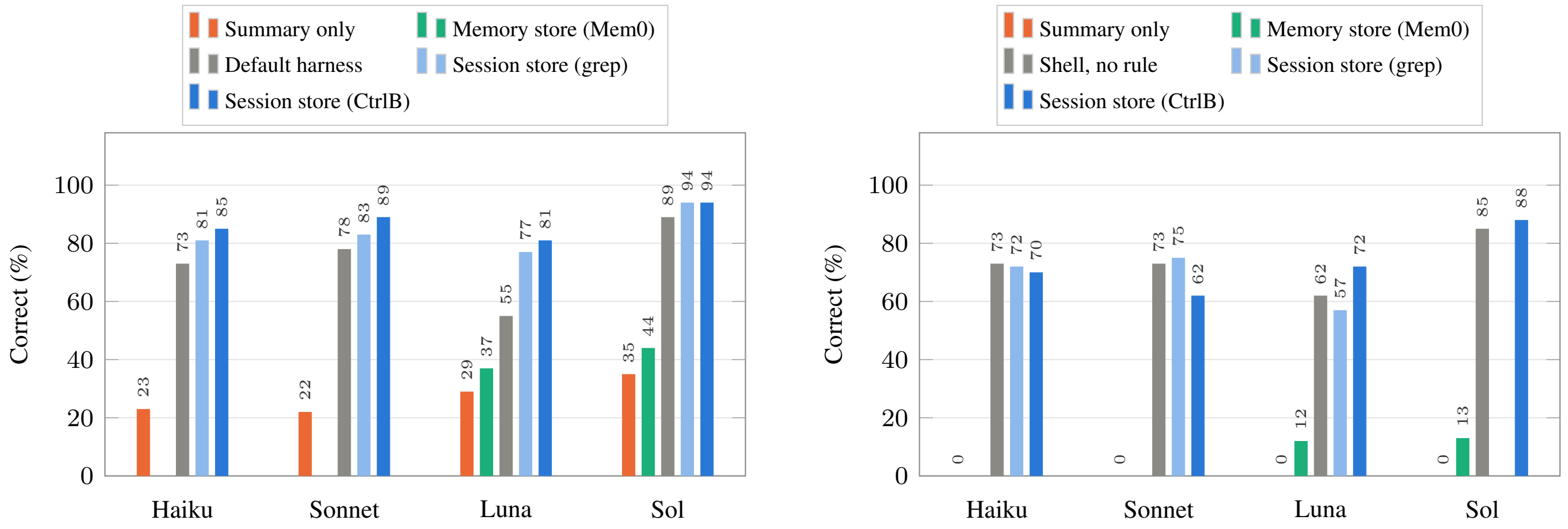
\begin{figure*}[!t]
\centering
\begin{minipage}[t]{0.49\textwidth}
\centering
\begin{tikzpicture}
\begin{axis}[designbars]
\addplot[fill=cSum, draw=none] coordinates {(Haiku,23) (Sonnet,22) (Luna,29) (Sol,35)};
\addlegendentry{Summary only}
\addplot[fill=cMem, draw=none] coordinates {(Luna,37) (Sol,44)};
\addlegendentry{Memory store (Mem0)}
\addplot[fill=cNone, draw=none] coordinates {(Haiku,73) (Sonnet,78) (Luna,55) (Sol,89)};
\addlegendentry{Default harness}
\addplot[fill=cGrepS, draw=none] coordinates {(Haiku,81) (Sonnet,83) (Luna,77) (Sol,94)};
\addlegendentry{Session store (grep)}
\addplot[fill=cRaw, draw=none] coordinates {(Haiku,85) (Sonnet,89) (Luna,81) (Sol,94)};
\addlegendentry{Session store (CtrlB)}
\end{axis}
\end{tikzpicture}

\small (a) Detail recall, 100 questions per family
\end{minipage}\hfill
\begin{minipage}[t]{0.49\textwidth}
\centering
\begin{tikzpicture}
\begin{axis}[designbars]
\addplot[fill=cSum, draw=none] coordinates {(Haiku,0) (Sonnet,0) (Luna,0) (Sol,0)};
\addlegendentry{Summary only}
\addplot[fill=cMem, draw=none] coordinates {(Luna,12) (Sol,13)};
\addlegendentry{Memory store (Mem0)}
\addplot[fill=cNone, draw=none] coordinates {(Haiku,73) (Sonnet,73) (Luna,62) (Sol,85)};
\addlegendentry{Shell, no rule}
\addplot[fill=cGrepS, draw=none] coordinates {(Haiku,72) (Sonnet,75) (Luna,57)};
\addlegendentry{Session store (grep)}
\addplot[fill=cRaw, draw=none] coordinates {(Haiku,70) (Sonnet,62) (Luna,72) (Sol,88)};
\addlegendentry{Session store (CtrlB)}
\end{axis}
\end{tikzpicture}

\small (b) Cross-session lookup, 60 questions
\end{minipage}
\caption{Memory designs with the same model. Haiku and Sonnet are Claude; Luna and Sol are Codex. With every model,
keeping the record searchable beats the compaction summary by 52--67 points on Detail recall, and beats a Mem0
store where we ran one. Absent bars denote unrun arms; zero labels denote measured zero accuracy. Mem0 was run on Codex only; Sol read Luna-written memories, and the Lookup store was
favourable. In (b), Haiku and Luna use the redesigned recall tool (Section~\ref{sec:lookupfix}); Sonnet uses the
first tool; Sol's CtrlB bar is the first tool without the rule, and Sol was not run with grep and the rule.
Claude arms judged by gpt-6-astra, Codex arms by gpt-6.1-sol.}
\label{fig:designs}
\end{figure*}

\subsection{Session store against memory store and compaction, across models}
\label{sec:designs}

\begin{table*}[!t]
\centering\small
\caption{Memory designs with the same model: share correct, and cost per question. Detail recall: 100 questions
per family (exact / gist in brackets). Lookup: 60 questions. Claude judged by gpt-6-astra, Codex by gpt-6.1-sol;
exact answers string-matched. --- = not run. $^a$Sol reading Luna-written memories. $^b$Redesigned recall tool.
$^c$First recall tool, no rule.}
\label{tab:designs}
\setlength{\tabcolsep}{3.6pt}
\begin{tabular}{@{}llcccccc@{}}
\toprule
 & Model & \shortstack{Summary\\only} & \shortstack{Memory store\\(Mem0)} & \shortstack{Default\\harness} & \shortstack{Session store\\(grep)} & \shortstack{Session store\\(CtrlB)} & \shortstack{CtrlB $-$ summary\\(95\% CI)} \\
\midrule
\multirow{4}{*}{Detail}
 & Haiku  & 23 (11 / 38) & ---          & 73 (91 / 51) & 81 (95 / 64) & 85 (91 / 78) & $+62$ ($+48$, $+74$) \\
 & Sonnet & 22 (9 / 38)  & ---          & 78 (95 / 58) & 83 (95 / 69) & 89 (95 / 82) & $+67$ ($+55$, $+77$) \\
 & Luna   & 29 (11 / 50) & 37 (13 / 65) & 55 (50 / 61) & 77 (81 / 72) & 81 (83 / 78) & $+52$ ($+39$, $+63$) \\
 & Sol    & 35 (13 / 61) & 44$^a$ (15 / 78) & 89 (91 / 87) & 94 (93 / 96) & 94 (93 / 96) & $+59$ ($+48$, $+68$) \\
\midrule
\multirow{4}{*}{Lookup}
 & Haiku  & 0 & ---    & 73 & 72 & 70$^b$ & $+70$ ($+58$, $+81$) \\
 & Sonnet & 0 & ---    & 73 & 75 & 62 & $+62$ ($+50$, $+73$) \\
 & Luna   & 0 & 12     & 62 & 57 & 72$^b$ & $+72$ ($+60$, $+83$) \\
 & Sol    & 0 & 13$^a$ & 85 & ---  & 88$^c$ & --- \\
\midrule
\multirow{4}{*}{\$ / question}
 & Haiku  & 0.003 & ---   & 0.005 & 0.006 & 0.005 & \\
 & Sonnet & 0.073 & ---   & 0.074 & 0.083 & 0.082 & \\
 & Luna   & 0.001 & 0.003 & 0.002 & 0.003 & 0.003 & \\
 & Sol    & 0.021 & 0.071 & 0.044 & 0.059 & 0.046 & \\
\bottomrule
\end{tabular}
\end{table*}

Figure~\ref{fig:designs} and Table~\ref{tab:designs} compare four models in two families, with the model fixed and
only the memory changed. These runs keep the summary in context, as harnesses do today, so they measure what the
record adds on top of compaction.
\begin{itemize}
  \item \textbf{Summary-only recall was limited.} With the summary alone, every model answered 22--35\% of Detail
        questions and none of the lookup questions. Sonnet scored within a point of Haiku (22\% vs 23\%), and Sol 6 points above Luna. The
        missing details are not in the summary.
  \item \textbf{The session store recovered them, in every model.} With the CtrlB store and the rule, Detail
        recall rose by 52 to 67 points over the summary in all four models; with grep over the transcript file,
        by 48 to 61. Most of the gain is on exact details ($+72$ to $+86$ points), but gist answers also rose by
        28 to 44 points. On lookup, the store with a small model found the next step and its outcome in 70--72\%
        of questions.
  \item \textbf{Mem0 improved gist recall more than exact recall.} Mem0 added 8 points over the summary for Luna
        and 9 for Sol on Detail recall, and 12--13 points on lookup with a store that held every answer's source
        window. It kept much of the gist (65--78\% of gist questions, against 78--96\% for the session
        store) but lost the strings: 13--15\% of exact details, against 83--93\% from the record. On the
        favourable lookup store, the 20 top-ranked memories contained the next command for 0 of 55 questions. The
        session store beat Mem0 by 44 points with Luna and 50 with Sol.
  \item \textbf{The default harness was in between.} With the transcript path and a shell, as Claude Code does,
        Claude recovers most exact details (91--95\%), but its gist answers stay 24--27 points below the CtrlB
        store (Section~\ref{sec:rule}).
\end{itemize}

\subsection{Smaller model with a session store against a larger model}
\label{sec:tier}

\begin{figure*}[!t]
\centering
\begin{minipage}[t]{0.49\textwidth}
\centering
\begin{tikzpicture}
\begin{axis}[diffs, height=5.6cm,
  ytick={1,2,3,4,5,6,7},
  yticklabels={{Sonnet: summary}, {Sonnet: default}, {Sonnet: store},
               {Sol: summary}, {Sol: Mem0}, {Sol: default}, {Sol: store}},
  xlabel={Small model with store minus large model\\(percentage points; 95\% CI)}, xlabel style={align=center},
  legend style={font=\scriptsize, draw=black!20, fill=white, at={(0.5,1.03)}, anchor=south, legend columns=2}, legend cell align=left]
\draw[black!55] (axis cs:0,0.4) -- (axis cs:0,7.6);
\draw[black!40, dashed] (axis cs:-5,0.4) -- (axis cs:-5,7.6);
\addplot+[est, mark=*, color=cRaw] coordinates {(63,1) += (11.4,0) -= (14.4,0) (7,2) += (8.9,0) -= (10.4,0)
  (-4,3) += (6.2,0) -= (7.9,0)};
\addlegendentry{Claude: Haiku}
\addplot+[est, mark=square*, color=cSum] coordinates {(46,4) += (11.3,0) -= (13.4,0) (37,5) += (9.0,0) -= (10.8,0)
  (-8,6) += (8.0,0) -= (7.3,0) (-13,7) += (5.5,0) -= (6.1,0)};
\addlegendentry{Codex: Luna}
\end{axis}
\end{tikzpicture}

\small (a) Detail recall (100 questions per family)
\end{minipage}\hfill
\begin{minipage}[t]{0.49\textwidth}
\centering
\begin{tikzpicture}
\begin{axis}[diffs, height=5.6cm,
  ytick={1,2,3,4,5,6,7},
  yticklabels={{Sonnet: summary}, {Sonnet: shell}, {Sonnet: store},
               {Sol: summary}, {Sol: Mem0}, {Sol: shell}, {Sol: store}},
  xlabel={Small model with store minus large model\\(percentage points; 95\% CI)}, xlabel style={align=center},
  legend style={font=\scriptsize, draw=black!20, fill=white, at={(0.5,1.03)}, anchor=south, legend columns=2}, legend cell align=left]
\draw[black!55] (axis cs:0,0.4) -- (axis cs:0,7.6);
\draw[black!40, dashed] (axis cs:-5,0.4) -- (axis cs:-5,7.6);
\addplot+[est, mark=*, color=cRaw] coordinates {(70,1) += (11.4,0) -= (11.7,0) (-3.3,2) += (11.9,0) -= (12.5,0)
  (8.3,3) += (12.8,0) -= (11.8,0)};
\addlegendentry{Claude: Haiku}
\addplot+[est, mark=square*, color=cSum] coordinates {(71.7,4) += (10.8,0) -= (12.1,0) (58.3,5) += (12.7,0) -= (13.1,0)
  (-13.3,6) += (13.3,0) -= (13.8,0) (-16.7,7) += (13.5,0) -= (13.8,0)};
\addlegendentry{Codex: Luna}
\end{axis}
\end{tikzpicture}

\small (b) Cross-session lookup (60 questions)
\end{minipage}
\caption{The small model (Haiku, Luna) with the CtrlB session store and the rule, against the large model of the
same family (Sonnet, Sol) with each alternative. Paired differences on the same questions; bars are 95\% intervals
resampling sessions. Solid line: no difference; dashed line: the $-5$-point non-inferiority margin. The small model
with a session store beats the large model with a summary or a memory store by 37--72 points in both families.
With a searchable record on both sides, Haiku's differences from Sonnet are inconclusive; Luna trails Sol. In (b) the small models use
the redesigned recall tool; Sonnet's session store uses the first tool with the rule, Sol's the first tool without
it.}
\label{fig:tier}
\end{figure*}
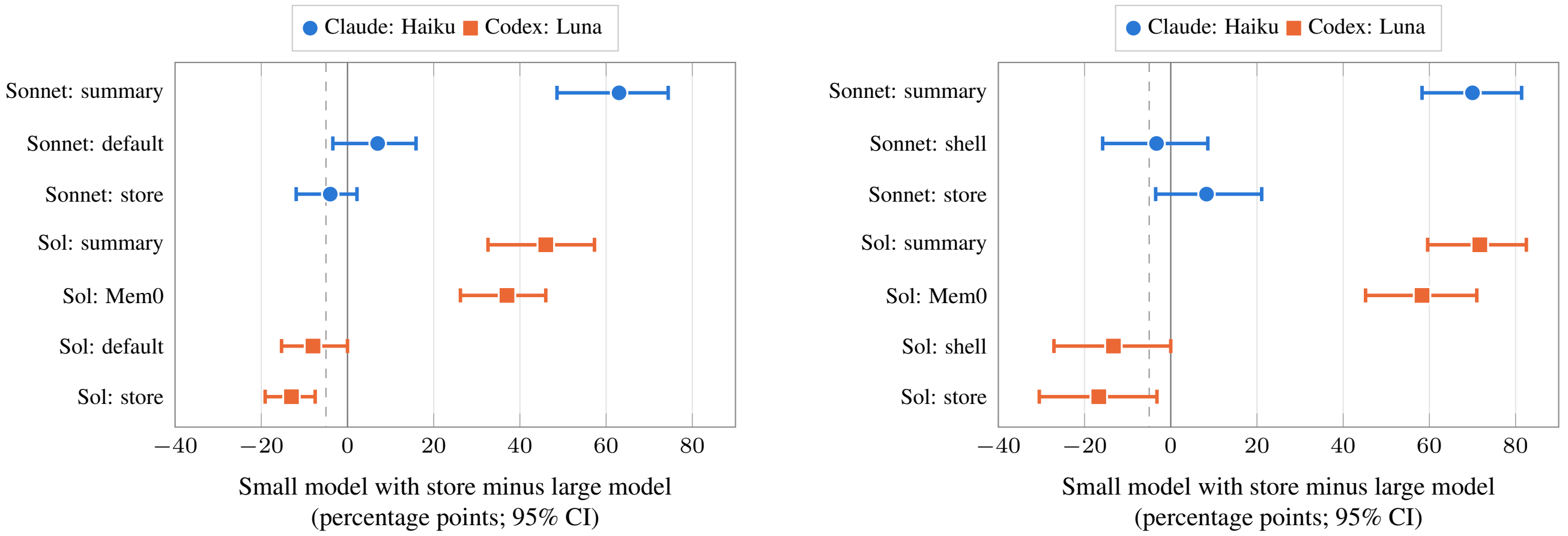

Figure~\ref{fig:tier} and Table~\ref{tab:tier} compare the small model with the CtrlB store against the large
model with each alternative.
\begin{itemize}
  \item \textbf{Against compaction, the small model won by a wide margin in both families.} Haiku with the
        store beat Sonnet with the summary by 63 points on Detail recall (CI $+49$ to $+74$) and 70 on lookup.
        Luna beat Sol by 46 (CI $+33$ to $+57$) and 72. The gap is largest on exact details ($+82$ for Haiku,
        $+70$ for Luna) and still positive on gist ($+40$ and $+17$).
  \item \textbf{Against a memory store, too.} Luna with the store beat Sol reading Mem0 memories by 37 points on
        Detail recall (CI $+26$ to $+46$) and 58 on lookup. On gist questions alone they tied (CI $-13$ to $+14$),
        so the difference is the exact details that Mem0 does not keep.
  \item \textbf{Against the large model with a searchable record, the families differed.} On Claude, Haiku's
        differences from Sonnet were not statistically significant. On Codex, Luna trailed Sol by 8 to 13 points
        on Detail recall and by 13 to 17 on lookup. The small models cost 1/8 to 1/26 as much per question
        (Table~\ref{tab:tier}; Appendix~\ref{app:cost}).
\end{itemize}

\begin{table}[t]
\centering\small
\caption{Small model with the CtrlB session store and rule, minus the large model with each alternative, Detail
recall (points, 95\% CI). Cost ratio: small model's cost per question over the large model's.}
\label{tab:tier}
\setlength{\tabcolsep}{3pt}
\begin{tabular}{@{}lcc@{}}
\toprule
Large model has & \shortstack{Claude\\Haiku $-$ Sonnet} & \shortstack{Codex\\Luna $-$ Sol} \\
\midrule
Summary only    & $+63$ ($+49$, $+74$) & $+46$ ($+33$, $+57$) \\
Memory store    & ---                  & $+37$ ($+26$, $+46$) \\
Default harness & $+7$ ($-3$, $+16$)   & $-8$ ($-15$, $0$) \\
Session store   & $-4$ ($-12$, $+2$)   & $-13$ ($-19$, $-8$) \\
\midrule
Cost ratio      & 1/14--1/15           & 1/8--1/26 \\
\bottomrule
\end{tabular}
\end{table}

\subsection{Why agents holding a summary skip the record}
\label{sec:rule}

\begin{table}[t]
\centering\small
\caption{The standing rule, Detail recall. Share of runs with no search of the record, and share correct, without
and with the rule. Claude: default harness (no rule) against the session store with grep (rule), same transcript
and shell. Codex: the same arm with and without the rule. Gain in points with 95\% CI.}
\label{tab:rule}
\setlength{\tabcolsep}{2.8pt}
\begin{tabular}{@{}llccc@{}}
\toprule
Model & Store & No search & Correct & Gain \\
\midrule
Haiku  & file  & 32\% $\to$ 6\%  & 73 $\to$ 81 & $+8$ ($0$, $+16$) \\
Sonnet & file  & 33\% $\to$ 18\% & 78 $\to$ 83 & $+5$ ($-2$, $+13$) \\
Luna   & file  & 60\% $\to$ 1\%  & 55 $\to$ 77 & $+22$ ($+12$, $+32$) \\
Luna   & CtrlB & 90\% $\to$ 0\%  & 34 $\to$ 81 & $+47$ ($+37$, $+57$) \\
Sol    & file  & 34\% $\to$ 0\%  & 89 $\to$ 94 & $+5$ ($0$, $+10$) \\
Sol    & CtrlB & 56\% $\to$ 0\%  & 72 $\to$ 94 & $+22$ ($+14$, $+31$) \\
\bottomrule
\end{tabular}
\end{table}

Claude Code already names the transcript in every summary. The agent with no summary searched every time
(Section~\ref{sec:nocompact}), but agents that hold a summary often do not, for two reasons.
\begin{itemize}
  \item \textbf{The agent does not know what the summary dropped.} With the default harness, Haiku and Sonnet
        searched the transcript for over 90\% of exact questions, where the summary visibly lacks the string, and
        scored 91--95\%. On gist questions the summary offers a plausible answer: Haiku (60\%) and Sonnet (62\%)
        answered without opening the transcript, and scored 51\% and 58\% on gist overall. Both skip equally
        often, so skipping is not a small-model habit on Claude.
  \item \textbf{On Codex, nothing points to the record.} None of the 32 compaction entries in our Codex sessions
        names the session file, so a stock Codex agent has only the summary. Even with the path or a recall tool, the
        agents rarely used it without the rule: Luna made no recall call in 90\% of runs and Sol in 56\%. Here
        skipping does track model size.
\end{itemize}
Table~\ref{tab:rule} shows the standing rule's effect. It nearly eliminated runs with no search. It raised Detail
recall by 22--47 points where skipping was common (Luna with either store, Sol with the recall tool), and by 5--8
points where it was not. On Claude the rule did less with the transcript file than with the CtrlB tool: gist
accuracy was 64--69\% when the agent grepped the raw JSONL transcript, against 78--82\% with the tool's rows.
The raw transcript has large JSON lines with escaped text; the tool returns individual blocks. Sonnet with the rule still skipped search on
24\% of gist questions. On Cross-session lookup, where the question says to search, the rule made no difference.

\subsection{Full context instead of a store}
\label{sec:fullctx}
A fourth option is a long window. On 50 of the Claude Detail questions we gave Haiku, which has a one-million-token
window, everything since the previous compaction (plus that summary) and no tools. It answered 56\% (41\% exact,
76\% gist), against 90\% for Haiku with the CtrlB store on the same questions ($+34$ points, CI $+12$ to $+53$).
Two things cost it. In 10 of the 29 exact questions the answer predates the previous compaction, so it was not in
the window. Where the answer was in the window, full context found 12 of 19 and the store 17 of 19: a long context
is not a reliable index~\cite{lostmiddle}. The median prompt was 186k tokens and the cost \$0.24 per question with
a cold cache, against \$0.005 for the store.

\subsection{Cross-session lookup and tool design}
\label{sec:lookupfix}

\begin{table}[t]
\centering\small
\caption{Cross-session lookup with the first and the redesigned recall tool, and with the transcript files and a
shell, all with the rule. Correct and exact-next values are percentages; Diagnosed: the 60 questions whose failure traces the redesign came from. Held-out: 60
others. Sonnet judge (Claude), Sol judge (Codex).}
\label{tab:lookupfix}
\setlength{\tabcolsep}{3pt}
\begin{tabular}{@{}llcccc@{}}
\toprule
Set & Arm & Correct & Exact next & Calls & s \\
\midrule
\multirow{3}{*}{Diagnosed} & Haiku, first tool & 60 & 73 & 5.5 & 17.6 \\
 & Haiku, redesigned & 77 & 89 & 3.6 & 14.2 \\
 & Haiku, files + shell & 77 & 80 & 6.2 & 28.3 \\
\midrule
\multirow{3}{*}{Held-out} & Haiku, first tool & 78 & 93 & 4.4 & 14.7 \\
 & Haiku, redesigned & 80 & 91 & 3.4 & 12.4 \\
 & Haiku, files + shell & 82 & 83 & 5.6 & 23.3 \\
\midrule
\multirow{2}{*}{Diagnosed} & Luna, first tool & 60 & --- & 3.9 & 11.2 \\
 & Luna, redesigned & 72 & --- & 2.8 & 8.9 \\
\bottomrule
\end{tabular}
\end{table}

Table~\ref{tab:lookupfix} gives the results. On the 60 lookup questions, Haiku with the first recall tool trailed
the shell by 17 points. Its traces showed three problems: search returned the oldest hits first with no paging, so
the failure in question was often past the cut; \emph{what followed} returned the next records, which were usually
text, not the next command; and replies were large and mixed sessions. The redesigned tool
(Section~\ref{sec:system}) raised Haiku from 60\% to 77\% ($+17$, CI $+5$ to $+29$; 70\% with the Astra judge),
level with the shell, with the next command quoted exactly 89\% of the time, fewer calls than either, and half the
shell's wall time. On 60 held-out questions it scored 80\%, against 78\% for the first tool and 82\% for the
shell. Luna rose from 60\% to 72\% ($+12$, CI $-2$ to $+25$). On held-out questions the redesign reduced calls and
latency, but its two-point accuracy gain was much smaller than on the diagnosed set.

\subsection{Speed}
\label{sec:speed}

Appendix~\ref{app:speed} (Figure~\ref{fig:speed}) reports request latency against context length, measured from the four engineers' real
sessions: on Claude, speed alone does not force an early compaction, and on Codex a shorter context is faster. Two
latencies matter for the store design. Compaction pauses the agent for a median of 61\,s
(Section~\ref{sec:cost}). Searching costs a few seconds per question: with the
store, Haiku took a median of 9.0\,s per Detail question against 4.2\,s from the summary alone, and 12--14\,s per
lookup question with the redesigned tool.

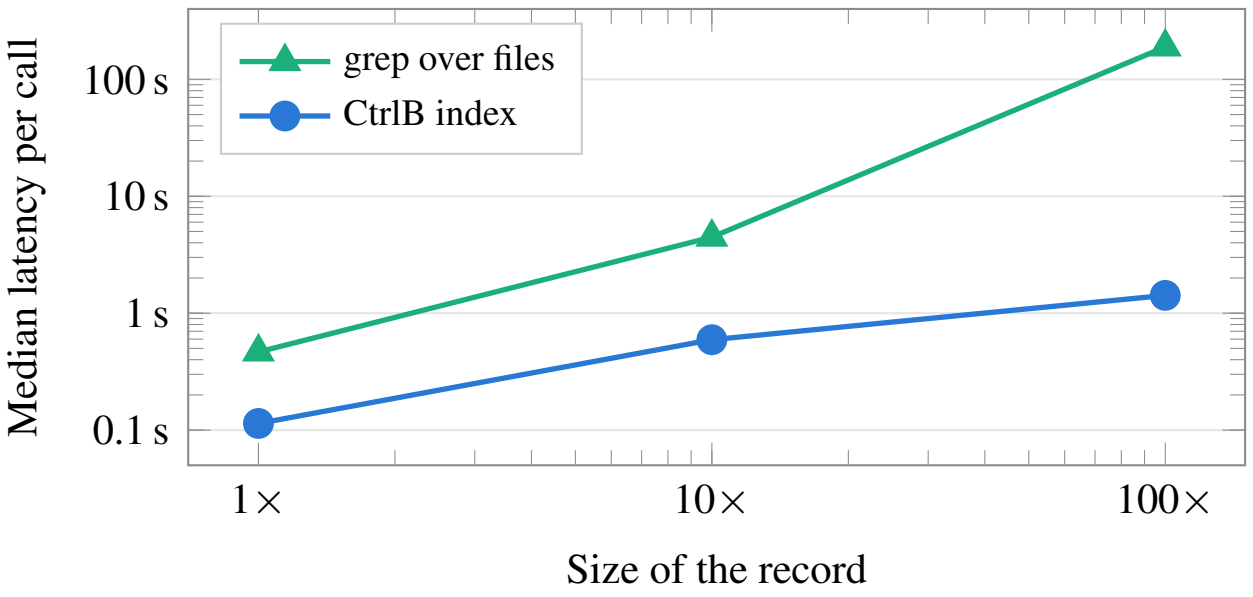
\begin{figure}[!htb]
\centering
\begin{tikzpicture}
\begin{loglogaxis}[width=\linewidth, height=4.6cm, xmin=0.7, xmax=150, ymin=50, ymax=400000,
  xtick={1,10,100}, xticklabels={1$\times$, 10$\times$, 100$\times$},
  ytick={100,1000,10000,100000}, yticklabels={0.1\,s,1\,s,10\,s,100\,s},
  xlabel={Size of the record}, ylabel={Median latency per call},
  ymajorgrids, grid style={black!10}, axis line style={black!45}, tick style={black!45},
  tick label style={font=\footnotesize}, label style={font=\footnotesize},
  legend style={font=\scriptsize, draw=black!20, fill=white, at={(0.03,0.97)}, anchor=north west}, legend cell align=left]
\addplot[color=cGrep, mark=triangle*, line width=1pt, mark size=2.6pt] coordinates {(1,466) (10,4472) (100,189376)};
\addlegendentry{grep over files}
\addplot[color=cRaw, mark=*, line width=1pt, mark size=2.4pt] coordinates {(1,114) (10,593) (100,1420)};
\addlegendentry{CtrlB index}
\end{loglogaxis}
\end{tikzpicture}
\caption{Recall-tool latency as the record grows: the median of the same nine whole-corpus calls at each size (the
only calls grep completed at 100$\times$). File sizes refer to normalised rows: approximately 0.36, 3.6 and 36\,GB, rather than the 2.24\,GB of raw source sessions.}
\label{fig:scale}
\end{figure}

\paragraph{Store latency at scale.}
The backends returned identical rows on the replayed calls. We next compared their latency as the record grew.
We replayed 120
whole-corpus search and what-followed calls from the lookup agents at 1, 10 and 100 times the corpus (copies of
every session with new ids: 1.7M and 17.0M rows, 3.6 and 36\,GB as files). CtrlB's median call took 147\,ms,
494\,ms and 1.05\,s (95th percentile 1.0, 2.0 and 3.6\,s). grep took 450\,ms and 4.1\,s at 1$\times$ and
10$\times$ (95th percentile 5.8 and 59\,s). At 100$\times$ the files exceed the machine's 31\,GB of memory, and each
of the nine grep calls that finished took 165--408\,s (Figure~\ref{fig:scale}). At 1$\times$ files are competitive,
and reading a range of one session file is faster from a file (3\,ms) than from the index (103\,ms). The index's
lead is on search, and it grows with the size of the record. Figure~\ref{fig:scale} shows only the nine calls that
grep completed at every scale, so replicated data and completion-selected calls limit how closely it predicts
production workloads.

\section{Discussion}
\label{sec:whynot}
\label{sec:discussion}

\paragraph{Short contexts and historical recall.}
Compaction-based short contexts kept agents accurate on the task benchmarks tested in~\cite{nvshort}. We suggest one
reason, as a hypothesis and not a measurement of their setting. What survives compaction is mostly the current
state, which acting next needs. What does not survive (commands, errors, numbers, earlier results) is what
questions about the past need, and a longer window supplied it less reliably than search. The store also serves questions about the team's
earlier sessions, which no in-session summary can answer. When several agents write to it under one run
identifier, it becomes a shared board for a swarm of agents, which a companion manuscript studies~\cite{swarms}.

\paragraph{Retaining a transcript is necessary but not sufficient.}
Claude Code names the transcript in every summary, yet Haiku and Sonnet answered 60--62\% of gist questions without
opening it. The default-harness arm differs from the CtrlB arm in both instruction and retrieval interface, and the
file-based ablation in Table~\ref{tab:rule} isolates the instruction more closely. Retention, retrieval interface
and instruction should be evaluated separately.

\paragraph{Lighter compaction, model size and cost.}
With a searchable record, compaction can shrink to a context reset: a fresh context with a pointer to the record
and an instruction to search it. Summaries may still matter for some tasks: some intervals allow losses, and
resumption plans benefited from orientation material such as a background note. A larger evaluation that executes
the resumed tasks is needed before treating these approaches as interchangeable. Small models with retrieval
outperformed larger models restricted to summaries or Mem0 on Detail recall. Non-significance does not establish
equivalence: Haiku's interval against Sonnet with the same store extends below the $-5$-point margin. These
results concern historical recall, not coding ability as a whole.


\section{Limitations}
\label{sec:limits}
\begin{itemize}
  \item \textbf{Four engineers, one company, six weeks.} All compaction points are in Claude Code sessions, so Codex
        models answer questions about Claude Code sessions. The memory-design and model-size comparisons use 100
        Detail and 60 lookup questions per family, so differences under about 10 points are usually inconclusive.
        The two families' Detail samples share 37 questions, so family comparisons are indicative.
  \item \textbf{Mem0 on Codex only, and favourable.} The Claude Mem0 arm was stopped for cost before its stores were
        built. The Codex lookup store held every answer's source window, and Sol read memories that Luna wrote.
        Other datasets and extraction configurations are untested.
  \item \textbf{Judges.} Model judges give the gist, lookup and resumption verdicts, with no human audit. A Codex
        model judges the Claude arms, which avoids within-family grading~\cite{llmjudge}. Sol judges the Codex arms,
        including answers it wrote; Astra agreed on 98\% of a 20\% sample. The full-context arm, the lookup
        redesign table and the first study use a Sonnet judge, which was more lenient than Astra on gist answers.
  \item \textbf{One compaction at a time.} Every question sits at a single compaction point. We did not measure how
        losses compound over a task that compacts many times, with or without the store.
  \item \textbf{Summary-free resets with three models.} We tested dropping the summary with Sonnet (first study, no
        rule) and with Haiku and Luna (with the rule), not with Sol or Astra. Resumption has 38 points, so its
        5--16-point differences in staying on task are near the limit of what it can resolve; the sign was the same
        in both families. Luna's intervals for recall without the summary reach $-10$, so non-inferiority at $-5$ is
        not established there. The running note's prompt was written once and not tuned. One Luna pointer-only run
        hit the tool-call cap and is counted as wrong.
  \item \textbf{Questions and runs.} We ran each question once; repeats in the first study moved an arm by up to 6
        points on 50 questions. Haiku phrased the exact questions and Sonnet wrote the gist questions from windows of
        the record, with the summary hidden from both. Their wording follows the record, which may help search arms.
  \item \textbf{Mixed tool versions on lookup.} Sonnet's and Sol's lookup store arms used the first recall tool, and
        the small models the redesigned one, which added 12--17 points on its diagnosed questions.
  \item \textbf{Recall and plans, not completed long tasks.} Resumption judges plans and does not run the tasks, and
        no experiment runs a long task to the end with and without the store. Short-context results on task
        benchmarks~\cite{nvshort} and our recall results are not yet joined in one experiment.
  \item \textbf{Observational speed; exact-token search only.} The speed figures come from the engineers' own
        sessions, where later calls may do different work, and request start times are approximated. We ran no
        controlled replay at different context lengths. We did not test paraphrase search (vector or fuzzy).
\end{itemize}

\section{Conclusion}
\label{sec:conclusion}
Short contexts kept long-horizon agents accurate in recent benchmarks~\cite{nvshort}. We asked what they lose about
their past. In this corpus, compaction preserved current-state answers far more often than exact details, and
agents restricted to the post-compaction context scored 22--35\% on Detail recall. On 50 questions, Haiku with the
history since the previous compaction scored 56\%, against 90\% with search (Section~\ref{sec:fullctx}). With the
full record searchable and an instruction to search it, recall reached 81--94\% in every model tested, and agents
answered questions about the team's earlier sessions that no summary answered. Removing the summary preserved
observed recall in the tested configurations at lower answer cost, although some intervals allow meaningful
losses. These recall results support a short context with access to the complete record and an instruction to
search it. Whether this improves completed long tasks remains open.

\paragraph{Data and ethics.}
The corpus is the authors' and their teammates' work sessions, shared by the teammates for this study and redacted
by the exporter. It contains proprietary code and is not released. This paper reports only aggregate results and
paraphrased question shapes.

\paragraph{Competing interest.}
The author works at CtrlB Explore Inc., which builds CtrlB Recall. Behind the same recall tool, a plain-file
\tool{grep} backend returned identical rows on the replayed calls.


\appendix
\sloppy
\section{Additional results}
\label{app:extra}

\subsection{First study: detailed comparison}
\label{app:first}

Table~\ref{tab:nocompact} gives the first-study results behind Figure~\ref{fig:detail} (Section~\ref{sec:nocompact}).

\begin{table*}[!t]
\centering\small
\caption{Detail recall with the record searchable, with and without the summary: Sonnet~5.5, 309 questions (174 exact, 135 gist) at 70
compaction points, no standing rule. \emph{Lost}: the 244 questions whose answer is not in the post-compaction
context. Times and costs are medians per question.}
\label{tab:nocompact}
\setlength{\tabcolsep}{4.5pt}
\begin{tabular}{@{}lcccccccc@{}}
\toprule
Arm & Exact & Gist & All (95\% CI) & Lost & s (95th pct.) & \$ & Tool calls & Gist runs with no tool call \\
\midrule
Kept by compaction (measured)  & 0.06 & 0.40 & 0.21 (0.16--0.29) & 0    & --          & --    & -- & -- \\
Default harness (compaction)   & 0.94 & 0.73 & 0.85 (0.82--0.88) & 0.83 & 9.8 (41.7)  & 0.067 & 2  & 62\% \\
Summary + recall (CtrlB)       & 0.94 & 0.88 & 0.92 (0.88--0.95) & 0.90 & 9.3 (20.1)  & 0.047 & 1  & 48\% \\
Summary + recall (grep)        & 0.95 & 0.93 & 0.94 (0.91--0.97) & 0.93 & 9.3 (20.6)  & 0.045 & 1  & 48\% \\
\textbf{Record only + recall} & 0.94 & 0.93 & 0.94 (0.92--0.96) & 0.93 & 11.0 (24.0) & \textbf{0.033} & 2 & \textbf{0\%} \\
\bottomrule
\end{tabular}
\end{table*}

\begin{itemize}
  \item \textbf{Recall.} The agent with no summary answered 0.94 of the 309 questions (0.94 exact, 0.93 gist). The
        two arms that had the summary and the same tool scored 0.92 and 0.94 (17 vs 10 discordant against the
        CtrlB arm, $p=0.25$). The stock harness scored 0.85, with gist 0.73 against 0.93 for the record-only arm
        (3 vs 31 discordant, $p<0.0001$). We found no recall advantage from adding the summary in this experiment.
  \item \textbf{Searching.} The record-only agent made a recall call in every gist run. With the summary in
        context, the same model and tool answered 48\% of gist questions without searching, and the stock harness
        62\%. This pattern is consistent with agents relying on plausible answers in the summary instead of
        checking the record. Section~\ref{sec:rule} shows the same skipping in every model.
  \item \textbf{Cost and time.} The record-only arm was the cheapest, at a median of \$0.033 per question against
        \$0.045--0.067, because it does not carry the 29k-character post-compaction context into every call. It took
        1.7\,s longer (11.0 vs 9.3\,s), the time of its extra search.
  \item \textbf{Backends.} Given the same tool, the CtrlB and grep backends returned identical replies. Agent
        answers nevertheless differed on 10 vs 2 questions ($p=0.04$). Repeated runs also varied, so this
        difference cannot be attributed to backend retrieval alone (the stock harness scored 0.88, 0.86 and 0.82
        on a 50-question subset; summary + recall 0.96, 0.98 and 0.96). On lookup, a shell (0.81) and the first
        recall tool on grep (0.78) and CtrlB (0.77) tied, and the tool quoted the next command verbatim more often
        (0.90 vs 0.79, $p=0.004$).
\end{itemize}
The same 309 questions with Codex and no rule, all with the summary in context, show the skipping at full scale.
Luna scored 51\% with the transcript file and 32\% with the recall tool, Sol 88\% and 65\%, and Astra 84\% with the
file. On the 188 lookup questions, where the question itself says to search, Sol with the recall tool scored 87\%
and Luna 65\%.

\subsection{Model and compaction costs}
\label{app:cost}

\paragraph{Compaction.}
To price compaction, we re-summarised the record before four points with Opus. We sent the same prompt twice, so
that the second call reads the prompt cache, as a live compaction does. A cached summary cost \$0.08--0.23
(126k--822k tokens) and took 31--76\,s, because writing 3--5k tokens of summary dominates.

\paragraph{Running notes.}
Haiku wrote the 72 points' notes in 156 updates for \$0.96 (\$0.013 per point, a median 12\,s per update), and Luna
wrote them for \$0.54 (\$0.008 per point). The input for each update is bounded by the 120k-character batch and the
previous note. Total note-writing cost still grows with the number of updates.

\paragraph{Small and large models.}
On Claude, Haiku's differences from Sonnet were not statistically significant: $+7$ points against Sonnet's default
harness (CI $-3$ to $+16$, which passes the $-5$ margin) and $-4$ against Sonnet with the same store ($-12$ to
$+2$); on lookup, $-3$ against Sonnet's shell and $+8$ against Sonnet's first-tool store. On Codex, Luna trailed Sol
by 8 points against Sol's default harness (CI $-15$ to $0$), by 13 against Sol with the same store ($-19$ to $-8$),
and by 13 to 17 on lookup. Astra, Codex's frontier model at five times Sol's price, scored 85\% with the default
harness, below Sol; Luna with the store was 4 points under Astra (CI $-13$ to $+4$).
Haiku with the store answered at \$0.005 per question, against \$0.074--0.082 for Sonnet. Luna's \$0.003 per
question is a sixteenth of Sol's \$0.044--0.046. Table~\ref{tab:tier} gives the cost ratios and
Table~\ref{tab:designs} the cost per question of every design.

\subsection{Observational speed analysis}
\label{app:speed}

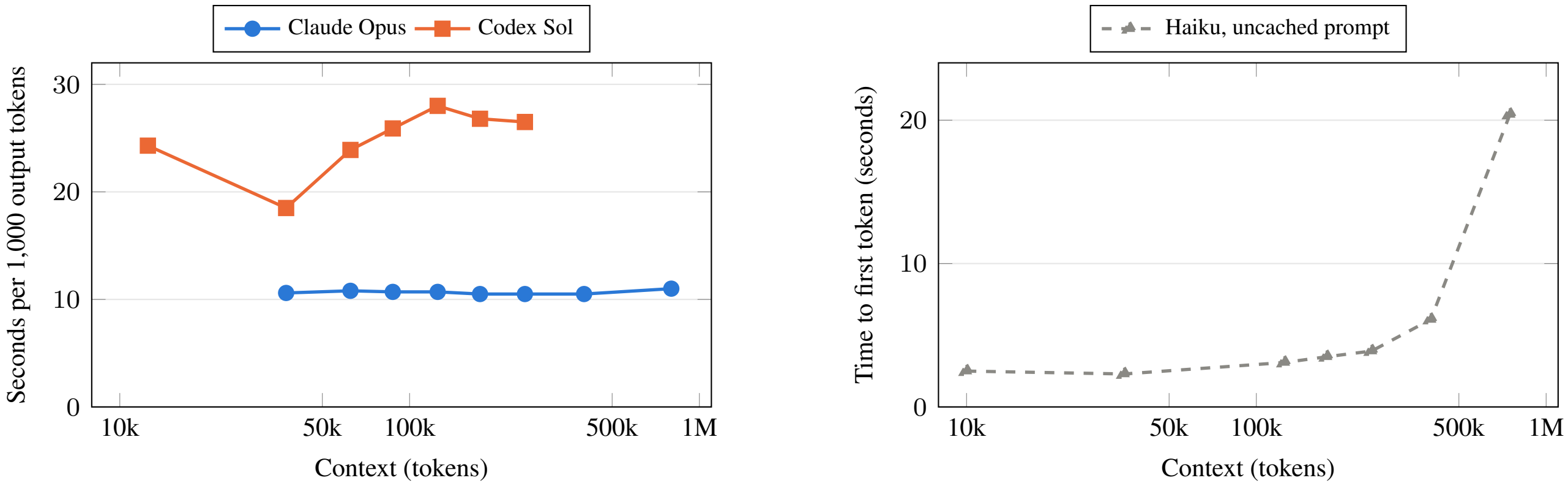
\begin{figure*}[!t]
\centering
\begin{minipage}[t]{0.485\textwidth}
\centering
\begin{tikzpicture}
\begin{semilogxaxis}[width=0.96\linewidth, height=5.3cm, xmin=8, xmax=1100, ymin=0, ymax=32,
  xtick={10,50,100,500,1000}, xticklabels={10k,50k,100k,500k,1M},
  xlabel={Context (tokens)}, ylabel={Seconds per 1,000 output tokens},
  ymajorgrids, grid style={black!10}, tick label style={font=\footnotesize}, label style={font=\footnotesize},
  legend style={font=\scriptsize, at={(0.5,1.03)}, anchor=south, legend columns=2}]
\addplot[color=cRaw, mark=*, line width=1pt] coordinates {(37.5,10.6) (62.5,10.8) (87.5,10.7)
  (125,10.7) (175,10.5) (250,10.5) (400,10.5) (800,11.0)};
\addlegendentry{Claude Opus}
\addplot[color=cSum, mark=square*, line width=1pt] coordinates {(12.5,24.3) (37.5,18.5) (62.5,23.9)
  (87.5,25.9) (125,28.0) (175,26.8) (250,26.5)};
\addlegendentry{Codex Sol}
\end{semilogxaxis}
\end{tikzpicture}

\small (a) Real-session calls: output-normalised latency
\end{minipage}\hfill
\begin{minipage}[t]{0.485\textwidth}
\centering
\begin{tikzpicture}
\begin{semilogxaxis}[width=0.96\linewidth, height=5.3cm, xmin=8, xmax=1100, ymin=0, ymax=24,
  xtick={10,50,100,500,1000}, xticklabels={10k,50k,100k,500k,1M},
  xlabel={Context (tokens)}, ylabel={Time to first token (seconds)},
  ymajorgrids, grid style={black!10}, tick label style={font=\footnotesize}, label style={font=\footnotesize},
  legend style={font=\scriptsize, at={(0.5,1.03)}, anchor=south}]
\addplot[color=cNone, mark=triangle*, line width=1pt, dashed] coordinates {(10,2.5) (35,2.3)
  (125,3.1) (175,3.5) (250,3.9) (400,6.1) (750,20.4)};
\addlegendentry{Haiku, uncached prompt}
\end{semilogxaxis}
\end{tikzpicture}

\small (b) Single cold calls: first-token latency
\end{minipage}
\caption{Latency against context length. (a) Median time per 1,000 output tokens, by context bucket, for
17,050 Opus~5.5 requests and 1,932 Sol calls in the engineers' sessions; only calls with at least 300 output
tokens are included. (b) Time to first token for single uncached Haiku calls from the full-context and
summary-only arms. The panels measure different quantities and use separate axes.}
\label{fig:speed}
\end{figure*}

The session files record a timestamp and token counts for every model call, so we measured request latency from the
four engineers' real sessions (Figure~\ref{fig:speed}).

\begin{itemize}
  \item \textbf{Claude: context barely slows a request.} Over 17,050 Opus~5.5 requests, median time per request
        rose from 3.6\,s at 25--50k tokens of context to 8.2\,s above 500k, but output per request rose with it
        (238 to 678 tokens), and the time per 1,000 output tokens stayed at 10.5--11.0\,s from 25k to 1M. A linear
        fit gives 0.14\,s more per request per 100k tokens of context at fixed output. Prompt caching reuses
        processing from earlier turns; these observational data do not isolate its causal effect.
  \item \textbf{Codex: Sol slows down.} Over 1,932 Sol calls, the median call took 4.9\,s at 25--50k tokens and
        10.0\,s at 100--150k, and the time per 1,000 output tokens rose from 18.5\,s to 26--28\,s above 75k. The
        fit gives 3.2\,s more per call per 100k tokens at fixed output.
  \item \textbf{Cold cache: long contexts are slow.} A single uncached Haiku call took 2.5\,s to its first token
        under 20k tokens and 3--4\,s up to 300k, then 6.1\,s at 300--500k and 20.4\,s at 500k--1M. After a pause
        long enough for the cache to expire, a long session pays this on its next turn.
\end{itemize}
On Claude, then, speed alone does not force an early compaction: a long cached context costs a fraction of a second
per request, and each compaction costs a minute. On Codex a shorter context is faster. With the store, that short
context can be a fresh one that points to the record, which is as short as a summary and needs no minute to write.

\subsection{Experimental details}
\label{app:details}

\paragraph{Mem0 store.}
Extraction calls go to gpt-6-luna through a local adapter. We build each Detail store by adding the record before
the point in windows (2,061 adds over 61 points). A full-team lookup store needs 10,468 adds; we stopped one at
1,224 for cost. The build of the lookup store took 47 minutes and \$3.56 at list prices. A Sol-written store was
estimated at \$83 and not built.

\paragraph{Judges.}
Astra judged 1,491 calls in the Claude comparison; it never ran a command and always gave a parseable verdict. It
agreed with an earlier Sonnet judge on 392 of 420 Haiku answers. It judged 639 calls on the Luna side of the
lighter-compaction runs, with the same prompts. Astra re-judged a seeded 20\% sample of the other Codex arms, with
agreement 420/428, 64/66 and 53/54 in the three Codex runs. Opus re-judged 20\% of the Sonnet-judged answers, with
93.8\% agreement.

\paragraph{File access.}
For example, Haiku with the CtrlB store over the default harness gained $+12$ points on all questions and $+13.6$
on the clean 88 after we dropped the questions with out-of-folder reads.

\paragraph{Detail samples.}
The Claude sample has 35 sessions, 55 exact and 45 gist questions. The Codex sample has 35 sessions, 54 exact and 46
gist questions. The two samples share 37 questions.

\subsection{Survival of details through compaction}
\label{app:survival}

Survival falls with distance: the post-compaction context stated the answer to 33\% of near questions, 19\% of mid
questions and 15\% of far questions. By kind, survival is 0\% for commands and error lines, 4\% for numbers, 16\% for
paths, 38\% for decisions, 36\% for reasons, 42\% for results and 60\% for current state.
\section{Prompts}
\label{app:prompts}
\small
\paragraph{Detail recall.} \emph{[post-compaction context, for the arms with a summary; or ``You are continuing
coding-agent session \{id\}. Its earlier turns are no longer in your context. Retrieve what you need from its
record with the recall tools as you go.'']} --- New message from the user: \emph{\{question\}} Exact: ``Answer from
this session's history. The question asks for a path, number, error line or command: give it exactly as it
appeared. Keep the answer short. If you cannot find it, say so.'' Gist: ``Answer from this session's history in one
to three sentences. If you cannot find it, say so.''

\paragraph{Standing rule.} Detail recall, file: ``Your earlier turns in this session are not all in your context:
the compaction summary leaves details out. The full record of this session up to the compaction is the file
\{path\}. Before answering anything about earlier work in this session, search that file. Do not say you cannot
find something until you have searched.'' CtrlB: the second sentence names the recall tools and the session id.
Mem0: ``Memories extracted from this session before the compaction are searchable with the memory tool
(search\_memories).'' Lookup: ``The team's earlier coding-agent sessions are not in your context. [Where they are.]
Before answering anything about earlier work, search them. Do not say you cannot find something until you have
searched.''

\balance
\paragraph{Lookup.} \emph{\{question\}} Shell: ``The session records of the team's coding agents
(Claude Code and Codex) are under \{dir\}, one JSONL file per session, in one directory per engineer.'' Recall:
``Use the recall tools to search the team's coding-agent session records.'' All: ``Quote the next command exactly as
it was run, then say whether it worked.''

\paragraph{Full context.} ``Below is this coding-agent session so far, oldest first. \emph{\{previous summary and
records since the previous compaction, each body capped at 30,000 characters\}} New message from the user:
\emph{\{question\}}'' followed by the exact or gist instruction above.

\paragraph{Running note.} Writer (Haiku or Luna, one call per part of at most 120k characters of new records,
tool results cut to 600 characters, tool calls to 400, messages to 2,000): ``You keep a running hand-off note for a
long coding-agent session, so that the agent can pick the work up after its context is cleared. The session's full
record is kept and searchable by record number (seq), so the note does not need to hold details: it should orient
the agent and point into the record. Update the note below with the new records. Write the whole updated note, at
most 500 words, in these sections: 1.~Task\ldots{} 2.~Current state\ldots{} 3.~Next step\ldots{} 4.~Key decisions
and results, each with the seq numbers of the records that show it\ldots{} 5.~Approaches tried and dropped\ldots{}
6.~Files, commands and services in play\ldots{} Do not copy long outputs; prefer a seq pointer to a detail.''
\emph{\{previous note\}} \emph{\{new records\}}. In the answering prompt the note is headed ``Hand-off note for this
session, kept by a small model while the session ran'', followed by the kept messages, re-attached files and ``If
you need specific details from earlier in this session, search its record (session\_id \{id\}) with the recall
tools.'' The rule for the arms without the summary reads ``Your earlier turns in this session are not all in your
context. The full record of this session up to now is searchable with the recall tools (search\_\allowbreak sessions,
fetch\_records; session\_id \{id\}). Before answering\ldots''

\paragraph{Resumption.} \emph{[post-compaction context, or nothing]} [Since then] \emph{\{records after the
summary\}} [User's latest message] \emph{\{message\}} ``Reply to the user's latest message as you would, but do not
run anything: the repositories have changed since, so this is a plan only. Write: 1.~Task\ldots{} 2.~Next actions:
the first three concrete actions you would take, with exact commands, file paths or tool calls. 3.~Facts relied
on\ldots''

\paragraph{Question writers.} Exact (Haiku): one question per gold string, specific enough that the gold is the
only right answer, never containing it. Gist (Sonnet): ``Write ONE question about this excerpt that the user or the
agent would need answered to carry on with this work later: a decision that was made and what it was, the outcome
or result of something that was tried, the current state of a piece of work, why something was done, or an
approach that was dropped and why\ldots{} do not ask for a single exact string\ldots''

\paragraph{Judges} (each ends ``Reply yes or no''; the Codex judges add ``Answer only from the text above. Do not
run any commands or read any files.''). Gist: ``Does the candidate give the reference answer's key facts, without
contradicting it? Extra detail is fine; saying it could not find the answer is a no.'' Survival: ``Does the context
state the reference answer's key facts, so that an agent could answer the question from the context alone?''
Lookup: ``Does the candidate describe the same next step and the same outcome as the truth?'' Resumption: the
first-action, on-task and contradiction questions of Section~\ref{sec:setup}, one call each.

\end{document}